\documentclass{article}
\usepackage{ijcai22}

\usepackage{times}
\usepackage{soul}
\usepackage{url}
\usepackage[hidelinks]{hyperref}
\usepackage[utf8]{inputenc}
\usepackage[small]{caption}
\usepackage{graphicx}
\usepackage{amsmath}
\usepackage{amsthm}
\usepackage{amssymb}
\usepackage{booktabs}
\usepackage{algorithm}
\usepackage{algorithmic}
\usepackage{multirow}
\usepackage{caption}
\title{ESGCN: Edge Squeeze Attention Graph Convolutional Network \\for Traffic Flow Forecasting }

\author{
Sangrok Lee
\and
Ha Young Kim
\affiliations
Yonsei University\\
\emails
lsrock1@yonsei.ac.kr,
hayoung.kim@yonsei.ac.kr
}

\begin{document}

\maketitle

\begin{abstract}
Traffic forecasting is a highly challenging task owing to the dynamical spatio-temporal dependencies of traffic flows. 
To handle this, we focus on modeling the spatio-temporal dynamics and propose a network termed Edge Squeeze Graph Convolutional Network (ESGCN) to forecast traffic flow in multiple regions.
ESGCN consists of two modules: W-module and ES module.
W-module is a fully node-wise convolutional network.
It encodes the time-series of each traffic region separately and decomposes the time-series at various scales to capture fine and coarse features. 
The ES module models the spatio-temporal dynamics using Graph Convolutional Network (GCN) and generates an Adaptive Adjacency Matrix (AAM) with temporal features.
To improve the accuracy of AAM, we introduce three key concepts. 
1) Using edge features to directly capture the spatio-temporal flow representation among regions.
2) Applying an edge attention mechanism to GCN to extract the AAM from the edge features.
Here, the attention mechanism can effectively determine important spatio-temporal adjacency relations.
3) Proposing a novel node contrastive loss to suppress obstructed connections and emphasize related connections.
Experimental results show that ESGCN achieves state-of-the-art performance by a large margin on four real-world datasets (PEMS03, 04, 07, and 08) with a low computational cost.
\end{abstract}


\section{Introduction}

Traffic flow forecasting is a core component of intelligent transportation systems.
It is essential for analyzing traffic situations and aims at predicting the future traffic flow of regions using historical traffic data.
However, this task is challenging because of the heterogeneity and dynamic spatio-temporal dependence of traffic data.

Traffic data can be modeled using Graph Neural Networks (GNNs).
In such networks, the regions are represented as nodes and flows between regions as edges.
Graph Convolutional Network (GCN) is a type of GNN commonly used to handle traffic flow forecasting tasks.
It can adequately leverage the graph structure and aggregate node information~\cite{wu2019graph,Wu2020ConnectingTD,Kong2020STGATSG,Song2020SpatialTemporalSG,Li2021SpatialTemporalFG}. 
Because the edges define the intensity of the adjacency matrix in a graph operation, accurate edge graph is an important factor in determining the performance of a GCN.
Recent studies focus on capturing the connection patterns via an adaptive adjacency matrix (AAM) that is found in the training process~\cite{NEURIPS2020_ce1aad92,wu2019graph}. 
In our study, we focus on enhancing the AAM with the following three distinctive features.

First, we propose an Edge Squeeze (ES) module that directly uses spatio-temporal flows with edge features to construct an AAM. 
Recent studies in GCN revealed that edge features are equally important as node features~\cite{Gong2019ExploitingEF,Chen2021EdgeFeaturedGA}. 
While the edge features can be used to simulate the traffic flows between regions, to the best of our knowledge this is the first study to build an adjacency matrix from the edge features in this task. 
Existing methods use node embeddings which cannot accurately reflect the relationship among nodes because the embedding vectors are fixed for inference and only represent spatial nodes.
Therefore, they cannot handle dynamic patterns occurred in the inference and are unable to accurately capture spatio-temporal features.
However, the ES module leverages the temporal features directly to construct the AAM and reflects the changes of temporal features in the inference.
ES module creates three-dimensional (3D) spatio-temporal correlations beyond the spatial-specific embeddings.

Second, we develop a novel edge attention mechanism.
We further explore the edge features with the attention mechanism to refine the AAM.
Because the edge features represent the adjacency relations, we apply the attention mechanism to activate meaningful edges and suppress the others of the adjacency matrix.
The edge attention exploits feature's channel information such as SENet~\cite{Hu2020SqueezeandExcitationN}, referred to as the squeeze attention.
Existing methods apply transformer attention mechanism~\cite{Vaswani2017AttentionIA} which presents high computational burden.
The channel attention mechanism provides relatively lower computation cost and can generate more refined AAM.

Third, we introduce a novel node contrastive loss.
Previous studies computed the similarities between the node embeddings that were trained with a forecasting objective function for the AAM. 
This method generated the adjacency matrix without an explicit objective for the shape of the graph, consequently inducing suboptimal performance.
To overcome this, we maximize the difference between related and unrelated nodes to facilitate separation of the forecasting relevant nodes from the residuals.
This aids in preventing the propagation of information on unrelated nodes through AAM.

Additionally, we propose a backbone network, W-module, to extract multi-scale temporal features for traffic flow forecasting. 
W-module is a fully convolutional network that consists of node-wise convolution to handle time-series features of each node separately. 
Owing to its non-autoregressive attributes and receptive field of convolution layers, the W-module can extract multiple levels of temporal features from shallow to deep layers and provide a hierarchical decomposition.
We combine the ES module with W-module and propose ESGCN to forecast traffic flow.
The main contributions of this study are summarized as follows:
\begin{itemize}

\item We propose an end-to-end framework (ESGCN) using two novel modules: ES module and W-module. ESGCN effectively learns hidden and dynamic spatio-temporal relationships using edge features.

\item We introduce an edge attention mechanism and node contrastive loss to construct an AAM which captures accurately the relationships among nodes.

\item We perform extensive experiments on four real-world datasets (PEMS03, 04, 07, and 08); the results shows that ESGCN achieves state-of-the-art performance by a large margin with a low computational cost.
\end{itemize}

\section{Related Work}


\indent \textbf{GCN for traffic forecasting}. GCN is a special type of convolutional neural network that is widely used in traffic forecasting tasks~\cite{hechtlinger2017generalization,Kipf2017SemiSupervisedCW}.
Recurrent-based GCN adopts a recurrent architecture, such as LSTM and replaces inner layers with GCN~\cite{li2017diffusion,bai2020adaptive}.
This type of GCN handles spatial and temporal features recurrently, however it has long-range memory loss problem.
STGCN~\cite{yu2017spatio}, GSTNet~\cite{fang2019gstnet}, and Graph WaveNet~\cite{wu2019graph} use fully convolutional architecture and graph operation.
They exploit spatial and temporal convolution separately to model spatio-temporal data.
These methods show relatively fast inference speed and improved performance on long-range temporal data.

\textbf{Attention mechanism for traffic forecasting}. Attention mechanisms are used to effectively capture spatio-temporal dynamics for traffic forecasting.
ASTGCN ~\cite{guo2019attention}, GMAN~\cite{Zheng2020GMANAG}, and STGRAT~\cite{Park2020STGRATAN} exploited the attention mechanism which considers changes in road speed and diverse influence of spatial and temporal network.
Existing approaches leveraged transformer attention~\cite{Vaswani2017AttentionIA} that computes key, query and value relations.
Contrastingly, our proposed network adopts a channel attention such as SENet~\cite{Hu2020SqueezeandExcitationN} and CBAM~\cite{Woo2018CBAMCB} which have relatively low computation cost and high speed.

\textbf{Adaptive Adjacency Matrix Construction}. Previous studies used a predefined adjacency matrix for GCN~\cite{li2017diffusion,yu2017spatio,Yao2019RevisitingSS,guo2019attention}. 
In a previous study, a spatial AAM was proposed as a supplementary for the predefined adjacency matrix for graph WaveNet~\cite{wu2019graph}.
STGAT~\cite{Kong2020STGATSG} also utilized the AAM, however, they were limited by spatial dependencies. 
AGCRN~\cite{bai2020adaptive} exploited an AAM solely based on spatial node embedding. 
In this study, we introduce a spatio-temporal based AAM that captures accurate relations.

\begin{figure*}[t]
\centering
\includegraphics[width=0.9\textwidth]{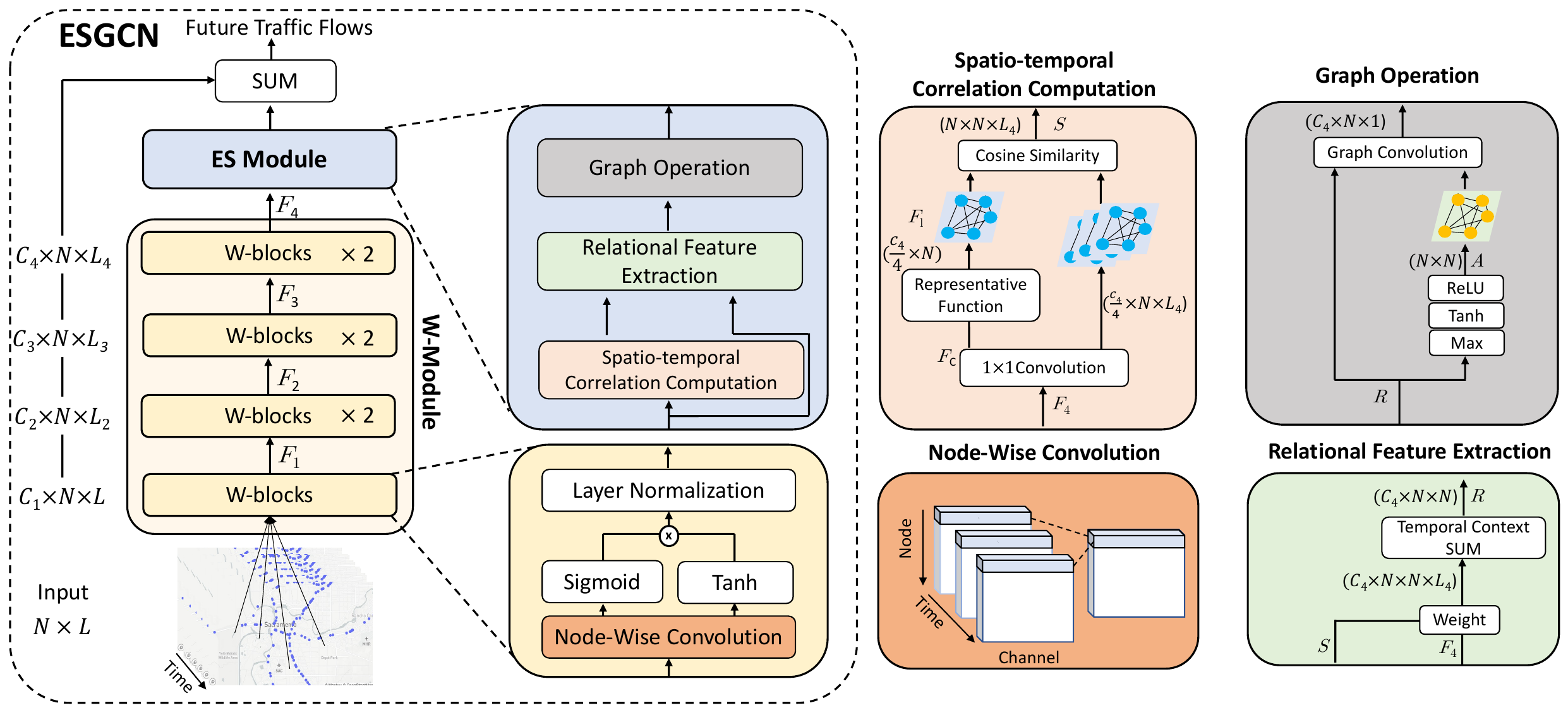} 
\caption{Overall architecture of ESGCN. 
ESGCN comprises the ES module and W-module consisting of four stages.
Each stage of the W-module consists of multiple W-blocks, namely gated node-wise convolution layers with a layer normalization. 
The ES module computes the spatio-temporal correlation, generates an adaptive adjacency matrix with attention mechanism, and finally conducts GCN.}
\label{fig:archi}
\end{figure*}

\section{Problem Definition}

To predict future traffic, univariate time-series data from each region ${\{X_1, X_2, X_3, \dots\}}$ are provided where $X_i\in \mathbb{R}^{n}$ is the traffic record at time step $i$ and $n$ is the number of regions.
The purpose is to find a function $\mathcal{E}$ that is capable of predicting the future of length $o$ by analyzing existing $T$-length past data. 
Traffic flows are generated for multiple regions and the problem is formulated as follows:
\begin{equation}
    \{X_{t+1}, \dots, X_{t+o}\} = \mathcal{E}(X_{t-T+1}, \dots, X_{t}),
\end{equation}
where $t$ is an arbitrary time step. 

\section{Proposed Method}
This section describes the edge squeeze graph convolution network (ESGCN).
The network comprises two modules: W-module and edge squeeze (ES) module, as shown in Fig. \ref{fig:archi}. 
W-module extracts region-specific temporal features and ES module produces an AAM using temporal features with an edge attention and 3D spatio-temporal relations.

\subsection{W-Module}
W-module extracts the time-series features of each node and is split into four W-block groups as illustrated in Fig. \ref{fig:archi}.
We call each group as stage.
To organize the W-module, we use a combination of gated node-wise convolutions (GNC) and a layer normalization~\cite{Ba2016LayerN}, namely W-block as the smallest unit.
GNC has two node-wise convolution layers: one for embedding features with a tanh function and the other for gating with a sigmoid function, as shown in Fig. \ref{fig:archi}.  
The two outputs are multiplied element-wise and summarized as follows:
\begin{equation}
    GNC(X) = sigmoid(\mathcal{C}(X)) \circ tanh (\mathcal{C}(X)), \label{equ:GCU}
\end{equation}
where $\mathcal{C}$ is a 2D convolution layer with a 1 $\times$ 3 kernel and 0$\times$1 padding, and $\circ$ is element-wise multiplication.
$X$ is an input feature of $X \in \mathbb{R}^{c \times n \times t}$, where $c$ is the number of channels, $n$ is the number of nodes, and $t$ is an arbitrary temporal dimension. 
This 1$\times$3 convolution layer is referred to as node-wise convolution. 
Given that the convolution kernel has a size of one in the node dimension and expands the receptive field in the time dimension, the W-block is only responsible for extracting temporal features. 
The output features of the GNC are fed to the layer normalization layer.
Each stage has 1, 2, 2, and 2 W-blocks with a stride of 1 for first stage and a stride of 2 for the others.
The output of $i^{th}$ stage is defined as $F_{i} \in \mathbb{R}^{~c_{i} \times n \times \frac{l}{2^{i-1}}}$, where $c_{i}$ is specified as a hyperparameter and $l$ is input time-series length. 

The receptive field of convolution layer expands as the feature passes through the various stages.
Therefore, the features of the early stages capture local signals and subtle changes and the features of late stages are learned for global signals and overall movements.
To take this advantages in handling multiple levels of signal size, the outputs of the intermediate stages are also used with the last output features for the final prediction.
In W-module, each stage has 3, 9, 12, and 12 receptive field sizes in a sequential order.
This can decompose the time-series hierarchically and each stage is efficiently trained to be responsible for possible signals.

\subsection{Edge Squeeze Module}
The ES module is a spatio-temporal feature extractor, which enhances temporal features with relations from the W-module. 
As described in Fig. \ref{fig:archi}, it reflects the node relations through three steps: spatio-temporal correlation computation, relational feature extraction, and GCN operation.

\textbf{Spatio-temporal correlation computation.} 
In this step, we construct spatio-temporal correlations to model flows among nodes.
The modeling involves three steps.
First, we feed the last features of W-module, $F_4$, to a single convolution layer to reduce the channel size and computation cost. This is defined as:
\begin{equation}
    F_c = \mathcal{C}(F_4), F_c \in \mathbb{R}^{~\frac{c_{4}}{4}\times n \times l_4},
\end{equation}
where $\mathcal{C}$ is $1 \times 1$ convolution of output channels and $l_4$ is $\frac{l}{8}$.

Second, we extract a temporal representative from $l_4$ temporal nodes in each region to compute spatio-temporal correlations between the representatives ($n$) and all nodes ($n \times l_4$).
The node is set in the last time step as a representative node.
This is inspired by the Markov decision process (MDP)~\cite{Bellman1961DynamicPA}, which considers only a state of current time step to decide an action at the subsequent time, and the representative of node is the closest value to the prediction value of the last element on the time dimension. 
The representative function $\mathcal{T}$ is as follows:
\begin{equation}
    F_l = \mathcal{T}(F_c), ~F_l \in \mathbb{R}^{~c_4 \times n},
\end{equation}
where $F_l$ denotes the last node on the time dimension.

Finally, the relations between the features and the representatives are computed.
The original feature, $F_c$ has $n \times l_4$ nodes, and representative, $F_l$ has $n$ nodes. 
The spatio-temporal correlations between $F_c$ and $F_l$ are in $n \times n \times l_4$ space. 
Inspired by previous studies \cite{wu2019graph,Li2021SpatialTemporalFG,Kong2020STGATSG}, a similarity function is adopted to measure the correlations between spatio-temporal nodes. 
The cosine similarity is adopted for this study. 
The result of cosine similarity is bounded from -1 to 1, and it is suitable for comparing high-dimensional features ~\cite{Luo2018CosineNU}. 
The cosine similarity $cos$ is defined as follows:
\begin{equation}
    cos(x_1, x_2) = \frac{x_1 \cdot x_2}{\Vert x_1 \Vert _2 \cdot \Vert x_2 \Vert _2},
\end{equation}
where $\cdot$ is dot product and $\Vert \cdot \Vert_2$ is Euclidean norm.
The final spatio-temporal correlations $S$ is defined as:
\begin{equation}
    S = cos(F_l, F_c), ~S \in \mathbb{R}^{n \times n \times l_4}
\end{equation}



\textbf{Relational features extraction.}
Based on the previous step, each node has $n \times l_4$ correlations.
We employ these correlations to weight $F_4$ and generate correlation-aware features, $\in \mathbb{R}^{c_4 \times n \times l_4}$, for each node.
Subsequently, we aggregate the temporal channel features, $l_4$, to reflect temporal contexts.
To facilitate computation, we expand the dimension of $S$ to $S^e \in \mathbb{R}^{n \times c_4 \times n \times l_4}$ (1 to $c_4$) and define $S_i^e \in \mathbb{R}^{c_4 \times n \times l_4}$ as the correlation of the $i^{th}$ node.
In following equations, $z(:,:,*)$ denotes $*^{th}$ tensor of $z$ on the last dimension.
The computation process is defined as:
\begin{equation}
    R(:,:,k) = \sum_{j\in l_4} S_k^e(:,:,j) \circ F_4(:,:,j),
\end{equation}
where $R(:, :, k) \in \mathbb{R}^{c_4 \times n}$ is the $k^{th}$ node's relational features between all nodes, $k = 1, 2, \dots, n$.

We use the relational features as a feature matrix for graph convolution.
This has an advantage over other features as it reflects temporal relations.
GCN conducts operation only considering of spatial node relations.
However, in spatio-temporal data, temporal relations need to be included.
In the relational features, each node has $n$ different nodes created by considering temporal contexts.
Thus, we use $R(:,:,k)$ tensor as $n$ neighbor nodes with $c_4$ dimension to predict $k^{th}$ node future flow.
It can reflect the temporal relations in the relational features and spatial relations in GCN operation.

\textbf{GCN operation.} GCN requires a feature matrix and an adjacency matrix.
As previously mentioned, the relational features are used as the feature matrix.
To construct the adjacency matrix, we apply an attention mechanism to the relational features.
Inspired by SENet~\cite{Hu2020SqueezeandExcitationN} and CBAM~\cite{Woo2018CBAMCB}, we adopt a channel attention mechanism known as squeeze attention to refine the AAM from the relational features.
The squeeze attention activates important spatial and channel positions.
Therefore, we use this mechanism to extract the edge positions' importance.
To squeeze the edge features, we feed the relational features to max operation, tanh, and ReLU activation as following:
\begin{equation}
    A = ReLU(tanh(max(R)))
\end{equation}
The outcome, $A \in n \times n$, is the generated AAM. 
The graph operation with $A$ and $R$ is defined as:
\begin{equation}
    F_g(:,k) = W \otimes R(:, :, k) \otimes A(k,:) + B,
\end{equation}
where $F_g(:,k) \in \mathbb{R}^{c_4}$ represents features for $k^{th}$ node, $W$ is a learnable weight denoted as $W \in \mathbb{R}^{c_4 \times c_4}$, $B$ is a bias as $B \in R^{c_4}$, and $\otimes$ is matrix multiplication. 
\subsection{ESGCN}

The proposed framework consists of the W-module and ES module.
The outputs of the first three stages of the W-module and ES module are employed to access all levels of features.
The computation process is summarized as follows:
\begin{equation}
    P = \sum_{i=1}^{t-1} \mathcal{C}^{~i}(F_{i}) +  \mathcal{C}^e(F_g),
\end{equation}
where $t$ is the number of stages, $\mathcal{C}^{~i}$ and  $\mathcal{C}^e$ are a $1 \times 1$ convolution layer for the $i$-th stage and ES module respectively.

To predict future flow, two fully connected layers with ReLU activation function are used. 
$P$ is fed to the two fully connected layers. 
\begin{equation}
    \hat{Y} = W_b \otimes ReLU (W_a \otimes P + B_a) + B_b, ~\hat{Y} \in \mathbb{R}^{~h \times n},
\end{equation}
where $W_a$ and $W_b$ are weights and $B_a$ and $B_b$ are bias of the fully connected layers and
$h$ is the number of time steps in the prediction.

\begin{figure}[t]
\centering
\includegraphics[width=0.9\columnwidth]{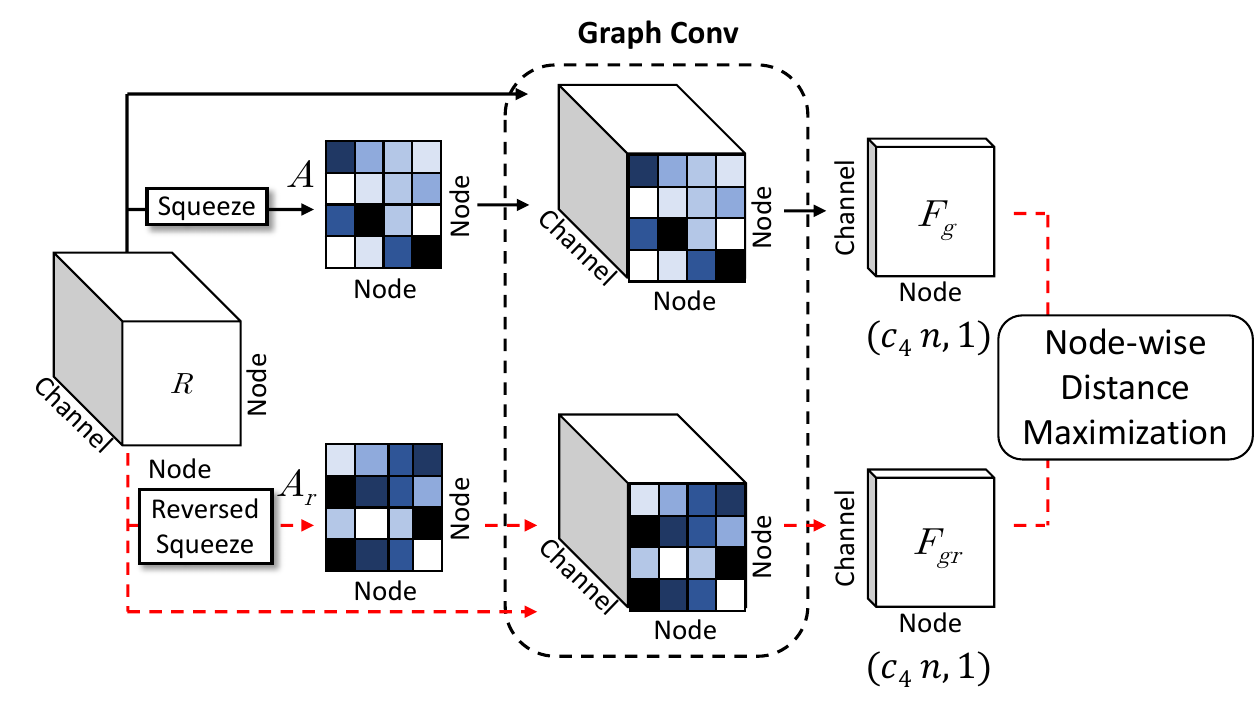} 
\caption{Proposed the node contrastive loss implementation. Note the branch with red dashed line is solely used for loss computation and not for inference.}
\label{fig:loss}
\end{figure}

\subsection{Loss Function}

We adopt the Huber loss and the proposed node contrastive loss.
Huber loss is defined as follows:

\begin{equation}
    h\big ( \hat{Y}, Y \big ) =\left\{
    \begin{aligned}
    &\frac{1}{2} (\hat{Y} - Y)^2, & \vert \hat{Y} - Y \vert < \delta \\
    &\delta \vert \hat{Y} - Y \vert - \frac{1}{2} \delta^2, & \vert \hat{Y} - Y \vert \geq \delta
    \end{aligned}
    \right. ,
\end{equation}
where $\delta$ is set to 1, $\hat{Y}$ denotes predicted values and $Y$ is ground truth values.
The objective function, $L_h$ is defined as follows:
\begin{equation}
L_h = \frac{1}{b \times o}\sum\limits_{k=1}^{b} \sum\limits_{i=1}^{o} h( \hat{Y}_{i}^k, Y_{i}^k ),
\end{equation}
where $\hat{Y}_i^k$ and $Y_i^k$ are the predicted value and ground truth of $i$-th time step of $k$-th sample in a mini batch respectively, and $b$ is the number of samples in a mini batch.

\textbf{Node contrastive loss}. This is used to effectively separate related and unrelated nodes.
As shown in Fig. \ref{fig:loss}, the reversed adjacency matrix, $A_r$, is generated.
We modify the squeeze function as follows:
\begin{equation}
    A_r = ReLU(-tanh(max(R)))
\end{equation}
Subsequently, the unrelated features are extracted using the graph operation in the reversed adjacency matrix.
Finally, we maximize the distance between the related features, $F_g$, with the original adjacency matrix and the unrelated features, $F_{gr}$, with the reversed adjacency matrix.
The node contrastive loss is defined as:
\begin{equation}
    L_n = \frac{1}{n} tr(F_g^{T} \otimes F_{gr}),
\end{equation}
where $tr$ is the trace of a matrix and $F_g^T \in \mathbb{R}^{~n \times c_4}$ denotes a transposed matrix. Minimizing an orthogonal of the multiplied matrix can be viewed as reducing the similarity between corresponding nodes based on a dot product. 

The final loss function is calculated as follows:
\begin{equation}
    L = L_h + \lambda L_n,
\end{equation}
where $\lambda$ is set to 0.1 in our experiments.

\begin{table*}[!htb]
	\centering
	\scalebox{0.80}{
		\begin{tabular}{c|c|cccccccc}
		\hline
			\multirow{2}{*}{\shortstack{Dataset}} &  \multirow{2}{*}{\shortstack{Metric}} & 
			\multirow{2}{*}{STGCN$\dagger$} &
			\multirow{2}{*}{ASTGCN$\dagger$} & 
			\multirow{2}{*}{\shortstack{Graph WaveNet$\dagger$}} &
			\multirow{2}{*}{GMAN*} &
			\multirow{2}{*}{STSGCN$\dagger$} & \multirow{2}{*}{\shortstack{AGCRN*}} & 
			\multirow{2}{*}{STFGNN$\dagger$} & \multirow{2}{*}{ESGCN} \\
			&&&&&&&&&\\ \hline
			
			\multicolumn{1}{c|}{\multirow{3}{*}{\shortstack{PEMS03}}} 
			& \multicolumn{1}{c|}{RMSE}
			& 30.12 $\pm$ 0.70 
			& 29.66 $\pm$ 1.68 & 32.94 $\pm$ 0.18 
			& 27.92 $\pm$ 1.15
			& 29.21 $\pm$ 0.56 & 28.19 $\pm$ 0.20 & 28.34$\pm$ 0.46 & \textbf{25.01 $\pm$ 0.34}\\
			
			\multicolumn{1}{c|}{}
			& \multicolumn{1}{c|}{MAE}
			& 17.49 $\pm$ 0.46 
			& 17.69 $\pm$ 1.43 & 19.85 $\pm$ 0.03 
			& 16.87 $\pm$ 0.31
			& 17.48 $\pm$ 0.15 & 15.93 $\pm$ 0.08 & 16.77 $\pm$ 0.09 & \textbf{15.05 $\pm$ 0.07}\\ 
			
			\multicolumn{1}{c|}{}
			& \multicolumn{1}{c|}{MAPE}
			& 17.15 $\pm$ 0.45 
			& 19.40 $\pm$ 2.24 & 19.31 $\pm$ 0.49
			& 18.23 $\pm$ 0.01
			& 16.78 $\pm$ 0.20 & 14.97 $\pm$ 0.17 & 16.30 $\pm$ 0.09 & \textbf{14.32 $\pm$ 0.01}\\ \hline
			
			\multicolumn{1}{c|}{\multirow{3}{*}{\shortstack{PEMS04}}} 
			& \multicolumn{1}{c|}{RMSE}
			& 35.55 $\pm$ 0.75 
			& 35.22 $\pm$ 1.90 & 39.70 $\pm$ 0.04
			& 30.71 $\pm$ 0.09
			& 33.65 $\pm$ 0.20 & 32.35 $\pm$ 0.22 & 31.88$\pm$ 0.14 & \textbf{30.38 $\pm$ 0.10}\\
			
			\multicolumn{1}{c|}{}
			& \multicolumn{1}{c|}{MAE}
			& 22.70 $\pm$ 0.64 
			& 22.93 $\pm$ 1.29 & 25.45 $\pm$ 0.03 
			& 19.16 $\pm$ 0.04
			& 21.19 $\pm$ 0.10 & 19.76 $\pm$ 0.09 & 19.83 $\pm$ 0.06 & \textbf{18.88 $\pm$ 0.05}\\ 
			
			\multicolumn{1}{c|}{}
			& \multicolumn{1}{c|}{MAPE}
			& 14.59 $\pm$ 0.21 
			& 16.56 $\pm$ 1.36 & 17.29 $\pm$ 0.24
			& 13.40 $\pm$ 0.01
			& 13.90 $\pm$ 0.05 & 13.13 $\pm$ 0.35 & 13.02 $\pm$ 0.05 & \textbf{12.70 $\pm$ 0.01}\\\hline
			
			\multicolumn{1}{c|}{\multirow{3}{*}{\shortstack{PEMS07}}}
			& \multicolumn{1}{c|}{RMSE}
			& 38.78 $\pm$ 0.58 
			& 42.57 $\pm$ 3.31 & 42.78 $\pm$ 0.07
			& -
			& 39.03 $\pm$ 0.27 & 35.09 $\pm$ 0.10 & 35.80 $\pm$ 0.18 & \textbf{33.93 $\pm$ 0.28}\\
			
			\multicolumn{1}{c|}{}
			& \multicolumn{1}{c|}{MAE}
			& 25.38 $\pm$ 0.49 
			& 28.05 $\pm$ 2.34 & 26.85 $\pm$ 0.05
			& -
			& 24.26 $\pm$ 0.14 & 21.17 $\pm$ 0.13 & 22.07 $\pm$ 0.11 & \textbf{20.55 $\pm$ 0.10}\\ 
			
			\multicolumn{1}{c|}{}
			& \multicolumn{1}{c|}{MAPE}
			& 11.08 $\pm$ 0.18 
			& 13.92 $\pm$ 1.65 & 12.12 $\pm$ 0.41
			& -
			& 10.21 $\pm$ 1.65 & 8.97 $\pm$ 0.07 & 9.21 $\pm$ 0.07 & \textbf{8.61 $\pm$ 0.01}\\\hline
			
			\multicolumn{1}{c|}{\multirow{3}{*}{\shortstack{PEMS08}}} 
			& \multicolumn{1}{c|}{RMSE}
			& 27.83 $\pm$ 0.20 
			& 28.16 $\pm$ 0.48 & 31.05 $\pm$ 0.07
			& 24.71 $\pm$ 0.13
			& 26.80 $\pm$ 0.18 & 25.69 $\pm$ 0.21 & 26.22 $\pm$ 0.15 & \textbf{24.59 $\pm$ 0.15}\\
			
			\multicolumn{1}{c|}{}
			& \multicolumn{1}{c|}{MAE}
			& 18.02 $\pm$ 0.14 
			& 18.61 $\pm$ 0.40 & 19.13 $\pm$ 0.08
			& 15.69 $\pm$ 0.02
			& 17.13 $\pm$ 0.09 & 16.22 $\pm$ 0.13 & 16.64 $\pm$ 0.09 & \textbf{15.56 $\pm$ 0.10}\\
			
			\multicolumn{1}{c|}{}
			& \multicolumn{1}{c|}{MAPE}
			& 11.40 $\pm$ 0.10 
			& 13.08 $\pm$ 1.00 & 12.68 $\pm$ 0.57
			& 10.04 $\pm$ 0.01
			& 10.96 $\pm$ 0.07 & 10.50 $\pm$ 0.20 & 10.60 $\pm$ 0.06 & \textbf{9.88 $\pm$ 0.08}\\\hline
			
			\multicolumn{1}{c|}{\multirow{3}{*}{\shortstack{Comp.\\ cost}}}
			& \multicolumn{1}{c|}{\# Param.}
			& 384,243 & 450,031 & 311,400
			& 229,569
			& 2,872,686 & 748,810 & 3,873,580 & \textbf{199,062}\\
			
			\multicolumn{1}{c|}{}
			& \multicolumn{1}{c|}{Train}
			& \textbf{9.14} & 31.79 & 29.30
			& 86.9
			& 43.16 & 21.57 & 43.02 & 26.65 \\ 
			
			\multicolumn{1}{c|}{}
			& \multicolumn{1}{c|}{Test}
			& 6.92 & 3.69 & \textbf{2.19}
			& 9.0
			& 48.45 & 2.43 & 48.45 & 3.02 \\\hline
			
	\end{tabular}}
	\caption{Performance and computation cost comparison of ESGCN and baseline models on PEMS03, PEMS04, PEMS07 and PEMS08 datasets. The values in bold indicates the best performances. The experiments are repeated ten times. MAPE metric is measured as a percentage.
	Comp. cost denotes computational cost. The Train is average time in seconds per epoch and the Test is total time of running test dataset in seconds. The computation costs are computed on the same machine with the same mini batch size on PEMS04 dataset.
	$\dagger$: Experimental results are excerpted from [Li and Zhu, 2021] *: Experiment is conducted by us. For all metrics, a smaller value denotes superior performance.}
	\label{tab:pems03-08}
\end{table*}

			
			
			
\section{Experiments}

\subsection{Implementation Details}

The proposed model is trained with an Adam optimizer ~\cite{Kingma2015AdamAM} for 50 epochs. 
The initial learning rate is 0.0003 and reduced by 0.7 every 5 epochs. 
The weight decay factor for L2 regulation is set to 0.0001, and the batch size is set to 30 for PEMS07 and 64 for PEMS03, 04, and 08. Training sessions are conducted on an NVIDIA Tesla V100 and Intel Xeon Gold 5120 CPU.

\begin{figure}[!t]
\centering
\includegraphics[width=1.0\columnwidth]{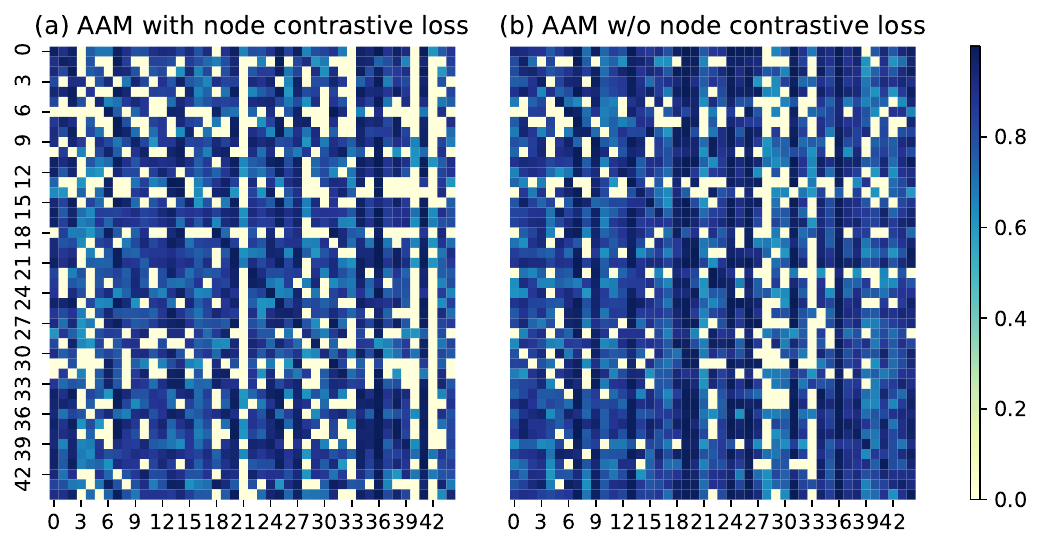} 
\caption{Heatmap of AAM with and without node contrastive loss on the first 45 nodes.}
\label{fig:aam}
\end{figure}

\subsection{Datasets}
\begin{table}[]
\centering
	\scalebox{0.8}{
	\begin{tabular}{c|cccc}\hline
		Datasets  & Nodes & MissingRatio & Range\\ \hline
		PEMS03    & 358 & 0.672\% & 9/1/2018 - 11/30/2018 \\
		PEMS04    & 307 & 3.182\% & 1/1/2018 - 2/28/2018 \\
		PEMS07    & 883 & 0.452\% & 5/1/2017 - 8/31/2017 \\
		PEMS08    & 170 & 0.696\% & 7/1/2016 - 8/31/2016 \\
		\hline
	\end{tabular}}
	\caption{Dataset description and statistics.}
	\label{tab:data_info}
\end{table}

The framework is validated on four real-world traffic datasets, namely: PEMS03, PEMS04, PEMS07, and PEMS08 ~\cite{Song2020SpatialTemporalSG}. 
Table \ref{tab:data_info} shows the description of each dataset. 
These four datasets contain generated traffic flows in four different regions of California using the Caltrans Performance Measurement System. 
Time-series data are collected at 5-minute intervals. 
Standard normalization and linear interpolation are used for stable training. 
For a fair comparison, all datasets are split into training, validation, and test data at a ratio of 6 : 2 : 2. Twelve time steps (1 h) are used to predict the next 12 time steps (1 h) and all experiments are repeated 10 times with random seeds. The test data performance is verified by selecting the model of the epoch that showed the best performance in the validation data.



\subsection{Baseline Methods}
ESGCN is compared with the following models on the same hyperparameters and official implementations:

\begin{itemize}

\item STGCN: Spatio-temporal graph convolutional networks, which comprise spatial and temporal dilated convolutions~\cite{yu2017spatio}.

\item ASTGCN: Attention-based spatial temporal graph convolutional networks, which adopt spatial and temporal attention into the model~\cite{guo2019attention}.

\item GraphWaveNet: Graph WaveNet exploits an adaptive adjacency graph and dilated 1D convolution~\cite{wu2019graph}.

\item GMAN: Graph multi-attention network uses spatial and temporal attention in graph neural network~\cite{Zheng2020GMANAG}.

\item STSGCN: Spatial–temporal synchronous graph convolutional networks, which utilize a spatio-temporal graph that extends the spatial graph to the temporal dimension~\cite{Song2020SpatialTemporalSG}.

\item AGCRN: Adaptive graph convolutional recurrent network for traffic forecasting. This model exploits node adaptive parameter learning and an adaptive graph~\cite{bai2020adaptive}.

\item STFGNN: Spatial–temporal fusion graph neural networks, which leverage fast-DTW to construct a spatiotemporal graph~\cite{Li2021SpatialTemporalFG}.

\end{itemize}


\begin{table*}[!t]
\centering
\scalebox{0.90}{
\begin{tabular}{c|c|c|c|c|c|ccc}
\hline
Case & ES module & Node contrastive loss & Attention Op.& $\lambda$ & Rep. function& RMSE & MAE & MAPE(\%) \\ \hline\hline
1    &    &    &  - &     -     & -    & 30.69&19.12 & 12.91   \\ \hline
2 & \checkmark & & Max & -    & Last &30.45 & 18.94 & 12.80 \\ \hline
3(ours) & \checkmark & \checkmark & Max & 0.1 & Last &\textbf{30.38} & \textbf{18.88} & \textbf{12.72} \\ \hline
4 & \checkmark & \checkmark & Max & 0.3 & Last &30.41 & 18.89 & 12.78 \\ \hline
5 & \checkmark & \checkmark & Max & 0.5 & Last &30.62 & 18.98 & 12.74 \\ \hline
6 & \checkmark & \checkmark & Max & 0.7 & Last &30.40 & 18.92 & 12.79 \\ \hline
7 & \checkmark & \checkmark & Max & 0.9 & Last &30.64 & 18.99 & 12.80 \\ \hline
8 & \checkmark & \checkmark & Avg & 0.1 & Last &30.46 & 18.98 & 12.91 \\ \hline
9 & \checkmark & \checkmark & Max + L & 0.1 & Last &30.48 & 18.89 & 12.87 \\ \hline
10 & \checkmark & \checkmark & Max & 0.1 & Middle & 30.55 & 19.04 & 12.80 \\ \hline
11 & \checkmark & \checkmark & Max & 0.1 & First & 30.64 & 19.12 & 12.87 \\ \hline
\end{tabular}}
\caption{Ablation experiments on the PEMS04 dataset. Attention Op. , $\lambda$, and Rep. function denote Attention Operation, $\lambda$ weight for node contrastive loss, and representative function. Max, Avg, Max + L denote Max operation, Average operation, and Max operation with learnable layers. 
Last, Middle, and Last are representative node positions.}
\label{tab:ablation_lambda}
\end{table*}
\subsection{Comparison with the Baseline Methods}

The proposed model is compared with state-of-the-art models.
ESGCN outperformed all other baselines in terms of RMSE, MAE, and MAPE, as shown in Table \ref{tab:pems03-08}. 
Compared to the current best performing models in each dataset (PEMS07: AGCRN, PEMS03,04, and 08: GMAN), ESGCN yields 4\%, 2.7\%, and 3.8\% relative improvements on average for all datasets in RMSE, MAE, and MAPE.
Graph WaveNet and AGCRN employ the AAM and STFGNN uses a non-adaptive spatio-temporal adjacency matrix.
Compared to these methods, the proposed model showed superior performance and guaranteed the effectiveness of our AAM which is refined with attention and node contrastive loss.
Based on the experimental results, ESGCN, has improved representation ability and exhibits promising forecasting performance.

\subsection{Computational Cost}
To evaluate the computational cost, we compare the number of parameters, the training time, and the inference time of our model with those of baselines in Table \ref{tab:pems03-08}.
ESGCN has the smallest parameters compared to other baselines.
In training time, ESGCN is faster than STFGNN and slightly slower than AGCRN.
ESGCN has the third-fastest inference speed.
Although AGCRN shows faster training and inference time, specifically, 5 and 0.6 s in training and inference, respectively, the differences are insignificant. 
Additionally, AGCRN requires three times more parameters than ours.
Especially, our model which has the channel attention shows faster running speed and includes fewer parameters than GMAN which leverages the transformer attention.
Considering its superior performance, ESGCN has an acceptable computational cost.

\subsection{Ablation Study}

\subsubsection{Components}
To validate the proposed ES module and the node contrastive loss, we conduct experiments (cases 1-3) as shown in Table \ref{tab:ablation_lambda}.
The model with only W-module (case 1) shows the lowest performance.
However, the W-module with the ES module which reflects traffic flows using GCN (case 2) achieves significant improvement.
ESGCN, consisting of the W-module, ES module, and the node contrastive loss (case 3) outperforms the others.
This highlights that the feature enhancement ability of ES module and the proposed loss function.

\subsubsection{Attention operation}
To extract important adjacency relations, our edge attention used max operation to squeeze channel features.
However, CBAM~\cite{Woo2018CBAMCB} also uses average operation to squeeze channel features.
In this section, we conduct an ablation study (cases 3, 8, and 9) on the attention operation in Table \ref{tab:ablation_lambda}.
We empirically discovered that the max operation is the most optimal setting.
Notably, the attention function with learnable layer (case 9) shows a lower performance because dimension reduction of the layer disturbs to extract the AAM~\cite{Wang2020ECANetEC}.

\subsubsection{Node contrastive loss}
We use a hyperparameter $\lambda$ to balance the Huber loss and node contrastive loss.
Although the node contrastive loss assists in the construction of the refined AAM, an excessive effect of this loss can negatively affect forecasting performance.
We empirically determine the magnitude of $\lambda$.
The experimental results (cases 3-7) are shown in Table \ref{tab:ablation_lambda}.
Based on the results, the optimal value of $\lambda$ is 0.1. 
We set $\lambda$ as 0.1 across in all the experiments.
In Fig \ref{fig:aam}, it shows additionally sparse AAM in which unrelated connections are removed.

\subsubsection{Representative function}
The representative function of ES module extracts temporal representatives on time dimension.
Inspired by MDP, the representative function returns the last step node of $F_c$.    
To test whether the last node can be a temporal representative, we conduct an experiment by replacing the last node with the first node, $F_c(:, :, 0)$, and the middle node, $F_c(:, :, l_4/2)$.
The experiment results (cases 3,10, and 11) are shown in Table \ref{tab:ablation_lambda}.
The closer the node to the initial time step is, the lower the performance.
Our representative function is based on the assumption the closest value to the prediction value can be the representatives.
This ablation study empirically shows the assumption is reasonable.
\section{Conclusion and Future Work}

This study proposes a novel method, ESGCN, to address traffic flow forecasting.
Experiments show that the proposed model achieves state-of-the-art performance on four real-world datasets and has the smallest parameters with relatively faster inference and training speed. 
In the future, given that ESGCN is designed as a general framework to handle spatio-temporal data, it can be applied to other applications that have spatio-temporal data structures such as regional housing market prediction and electricity demand forecasting. 
\newpage
\bibliographystyle{named}
\bibliography{ijcai22}

\end{document}